\documentclass[runningheads]{llncs}

\usepackage{eccv}

\usepackage{eccvabbrv}

\usepackage{graphicx}
\usepackage{booktabs}
\usepackage{wrapfig}
\usepackage{tabularx}

\usepackage[accsupp]{axessibility} 

\usepackage{multirow}
\usepackage{array}
\newcolumntype{L}[1]{>{\raggedright\arraybackslash}p{#1}}
\newcolumntype{C}[1]{>{\centering\arraybackslash}p{#1}}

\usepackage[colorlinks,citecolor=eccvblue]{hyperref}

\usepackage{orcidlink}

\begin{document}

\title{Active View Selection with Perturbed Gaussian Ensemble for Tomographic Reconstruction} 

\titlerunning{Perturbed Gaussian Ensemble}

\author{Yulun Wu\inst{1}\orcidlink{0009-0009-4982-9630} \and
Ruyi Zha\inst{2}\orcidlink{0009-0005-0410-1807} \and
Wei Cao\inst{1}\orcidlink{0009-0005-5163-6484} \and\\
Yingying Li\inst{1}\orcidlink{0000-0002-1858-4257} \and
Yuanhao Cai\inst{3,}\thanks{Corresponding author.}\orcidlink{0000-0002-8266-7102} \and
Yaoyao Liu\inst{1}\orcidlink{0000-0002-5316-3028}
}

\authorrunning{Y.~Wu et al.}

\institute{University of Illinois Urbana-Champaign \and
Australian National University \and
Johns Hopkins University \\
\email{\{yulun5, weicao3, yl101, lyy\}@illinois.edu\\ ruyi.zha@anu.edu.au\quad ycai51@jh.edu}
}

\maketitle

\begin{abstract}
  Sparse-view computed tomography (CT) is critical for reducing radiation exposure to patients. Recent advances in radiative 3D Gaussian Splatting (3DGS) have enabled fast and accurate sparse-view  CT reconstruction. Despite these algorithmic advancements, practical reconstruction fidelity remains fundamentally bounded by the quality of the captured data, raising the crucial yet underexplored problem of X-ray active view selection. Existing active view selection methods are primarily designed for natural-light scenes and fail to capture the unique geometric ambiguities and physical attenuation properties inherent in X-ray imaging. In this paper, we present \emph{Perturbed Gaussian Ensemble}, an active view selection framework that integrates uncertainty modeling with sequential decision-making, tailored for X-ray Gaussian Splatting. Specifically, we identify low-density Gaussian primitives that are likely to be uncertain and apply stochastic density scaling to construct an ensemble of plausible Gaussian density fields. For each candidate projection, we measure the structural variance of the ensemble predictions and select the one with the highest variance as the next best view. Extensive experimental results on arbitrary-trajectory CT benchmarks demonstrate that our density-guided perturbation strategy effectively eliminates geometric artifacts and consistently outperforms existing baselines in progressive tomographic reconstruction under unified view selection protocols.
\end{abstract}

\section{Introduction}
\label{sec:intro}
\begin{figure}[t]
  \centering
  \includegraphics[width=\textwidth]{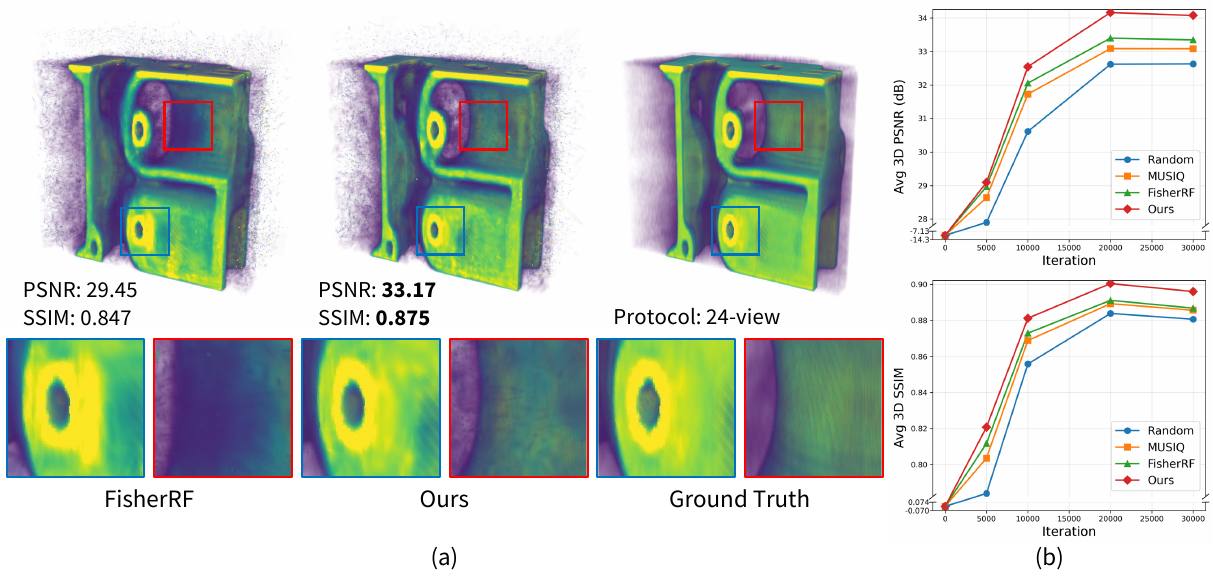}
  \caption{\textbf{(a)} We compare our approach against the state-of-the-art 3DGS-based active view selection method, FisherRF~\cite{fisherrf}, evaluating both quantitative metrics (3D PSNR$\uparrow$ and SSIM$\uparrow$) and qualitative visual fidelity, including zoomed-in details. Our method achieves the highest reconstruction quality and best preserves fine structural details. \textbf{(b)} We plot the training iterations \textit{vs.} the average 3D PSNR$\uparrow$ (dB) and SSIM$\uparrow$ on the synthetic dataset. As demonstrated, our proposed approach consistently delivers superior reconstruction fidelity throughout the progressive reconstruction process.} 
  \label{fig:teaser}
\end{figure}
X-ray computed tomography (CT) is an indispensable non-invasive imaging modality widely utilized in medical diagnosis, industrial inspection, and scientific research. During a CT scan, an X-ray machine captures multi-angle 2D projections that measure ray attenuation through the material. Tomographic reconstruction can then be performed to recover high-fidelity 3D anatomical structures from the 2D projections. However, the ionizing radiation associated with prolonged X-ray exposure poses significant health risks. 

To mitigate these risks, sparse-view CT has emerged as a critical scanning paradigm, aiming to maximize the reconstruction quality with a limited number of projection angles. Mathematically, this reduction in data acquisition transforms the tomographic reconstruction into a highly ill-posed inverse problem. Existing algorithms and solvers often struggle under such sparse conditions.

Recent breakthroughs in neural rendering, particularly 3D Gaussian Splatting (3DGS)~\cite{3dgs}, have demonstrated a remarkable performance superiority in both sparse-view novel view synthesis~\cite{fsgs,dngaussian,corgs} and surface reconstruction~\cite{neusurf,sparis,fatesgs,cao2026freeorbit4d}. Building on this, several methods~\cite{x_gaussian,r2gaussian,x2_gaussian} have successfully adapted 3DGS to X-ray imaging, enabling rapid and highly accurate volumetric reconstructions. However, despite these algorithmic leaps, practical reconstruction quality remains fundamentally bottlenecked by the captured data. Consequently, how to accurately select scanning viewpoints under a limited view-budget, so as to ensure both global silhouette coverage and the capture of local structural details, has emerged as a critical challenge in sparse-view tomography.

While crucial, active view selection for X-ray Gaussian Splatting remains largely unexplored. Existing view selection strategies~\cite{activenerf,fisherrf} are predominantly designed for natural-light scenes, relying heavily on surface occlusions and view-dependent specularities to estimate uncertainty or information gain. In stark contrast, X-ray imaging operates on the Beer-Lambert law~\cite{kak2001principles}, where the projection is a purely linear integral of the continuous density field along the ray path, without any occlusion. Furthermore, X-ray attenuation is inherently isotropic, meaning the constituent Gaussian primitives have no spherical harmonic (SH) parameters. Consequently, methods that depend on view-dependent color gradients and assume sparse, occlusion-bounded ray interactions fail to accurately capture the volumetric ambiguity in CT. As a result, these methods struggle to distinguish between the stretched artifacts and true high-density structures, often yielding redundant view selections that fail to provide the orthogonal constraints necessary to eliminate such artifacts.

In this study, we present \emph{Perturbed Gaussian Ensemble},\footnote[1]{Project page: \href{https://perturbed-gaussian-ensemble.cvmlgroup.web.illinois.edu/}{\texttt{perturbed-gaussian-ensemble.cvmlgroup.web.illinois.edu}}} a novel active view selection framework for progressive reconstruction specifically tailored to X-ray Gaussian Splatting and designed for flexible C-arm CT systems. Our core intuition is that under sparse-view constraints, geometric ambiguities typically manifest as fragile structures, such as uncertain boundaries and needle-like artifacts, whose projections vary dramatically when observed from informative, previously unseen angles. A valid next best view is the one that maximizes the exposure of this underlying structural instability.

To operationalize this insight, we utilize low-density Gaussian primitives as a proxy for uncertainty-prone regions, as they typically correspond to under-constrained boundaries, background noise, or degenerated artifact tails. By applying stochastic perturbations to the densities of these specific primitives, we construct an ensemble of plausible Gaussian density fields. For each candidate viewpoint, we render predictions across this ensemble and quantify their structural disagreement using the variance of the Structural Similarity Index Measure (SSIM)~\cite{ssim}. A high structural variance indicates that minor perturbations in uncertain regions provoke substantial structural discrepancies in that specific projection, marking the viewpoint as highly informative for resolving existing ambiguities. We therefore select the viewpoint that maximizes this structural variance as the optimal next best view.

We comprehensively evaluate our method against 2D-based paradigms, 3D-based paradigms, and conventional rule-based heuristics under two different view selection protocols within a hemispherical scanning search space. Experimental results demonstrate that our density-guided perturbation strategy effectively eliminates geometric artifacts and consistently outperforms existing baselines in progressive tomographic reconstruction, as illustrated in \cref{fig:teaser}.

Our contribution can be summarized as follows:
\begin{itemize}
    \item We propose a novel active view selection and progressive reconstruction framework designed for X-ray Gaussian Splatting. Our work bridges the gap between active learning and explicit radiative fields, addressing the unique physical and geometric challenges of sparse-view computed tomography.
    \item We introduce a novel uncertainty quantification strategy based on the \emph{Perturbed Gaussian Ensemble}. By applying stochastic density perturbations to low-density primitives that are highly susceptible to geometric degradation and measuring the structural disagreement in projection space, our method accurately localizes epistemic uncertainty and predicts the next best view.
    \item We establish an active view selection and progressive reconstruction benchmark for radiative Gaussian Splatting by adapting and evaluating previous state-of-the-art baselines. Extensive experiments demonstrate that our approach consistently outperforms existing paradigms, achieving superior volumetric reconstruction and novel view synthesis quality.
\end{itemize}

\section{Related Work}
\label{sec:related_work}

\subsubsection{Tomographic Reconstruction} aims to recover the 3D internal density field of an object from 2D X-ray projections. Traditional analytical methods~\cite{fdk,yu2006region} suffer from severe streak artifacts and noise amplification under sparse-view conditions. Iterative reconstruction algorithms~\cite{sart,manglos1995transmission, sauer2002local,sidky2008image} rely on optimization over iterations; however, they are computationally expensive and tend to over-smooth fine structural details. While supervised deep learning approaches~\cite{anirudh2018lose,ghani2018deep,chung2023solving,lee2023improving,liu2023dolce,liu2020tomogan,jin2017deep,ying2019x2ct,adler2018learned,lin2023learning,lin2024c} leverage semantic priors to enhance image quality, their generalization capabilities often remain constrained. NeRF-based frameworks~\cite{intratomo,naf,sax_nerf,ruckert2022neat,shen2022nerp} model the continuous density field using coordinate-based MLPs optimized purely by photometric losses from sparse projections. Despite achieving high-fidelity reconstructions, their training and rendering processes are prohibitively slow. The emergence of 3D Gaussian Splatting (3DGS)~\cite{3dgs} offers a compelling alternative due to its explicit parameterization and highly parallelized rasterization. Early adaptations to X-ray imaging~\cite{x_gaussian,ddgsct} focus primarily on novel view synthesis rather than direct volume retrieval. R$^2$-Gaussian~\cite{r2gaussian} achieves a critical breakthrough by identifying and rectifying the omission of a covariance-related scaling factor during 3D-to-2D projection. By introducing tailored radiative Gaussian kernels and a differentiable voxelizer, it enables direct, bias-free, and rapid static tomographic reconstruction. Subsequent works~\cite{li20253dgr,x2_gaussian,zhang2025x,liu20254drgs} further explore the sparse-view CT reconstruction paradigm or extend the radiative splatting framework into the temporal domain. However, active view selection for CT remains largely unexplored. Therefore, this study undertakes a more in-depth investigation and introduces a targeted solution to this problem.

\subsubsection{Active View Selection (AVS)} or next best view (NBV) planning aims to incrementally determine the most informative viewpoints for scene reconstruction to minimize data acquisition costs. It is closely related to continual learning and online learning~\cite{liu2020mnemonics,liu2021adaptive,liu2021rmm,liu2023online,luo2023class,liu2023continual,zhang2023continual,liu2024wakening,fischer2024inemo,duan2023prompt,zhu2025teachlmm,li2020online,li2019online,li2021online}. It originates from robotics research~\cite{connolly1985determination,scott2003view} and sees extensive exploration in 3D reconstruction~\cite{neunbv,gennbv,naruto,activegamer,prednbv,ran2023neurar}. With the rise of neural rendering, recent works adapt AVS to neural radiance fields~(NeRF)~\cite{nerf}. NeRF-based methods~\cite{activenerf,xiao2024nerfdirector,nvf,bayesrays,sunderhauf2023density} quantify uncertainty via variance estimation to guide view acquisition, but they remain computationally intensive and primarily apply to synthetic settings. The advent of 3D Gaussian Splatting~(3DGS)~\cite{3dgs} significantly accelerates active view selection and reconstruction due to its explicit representation and highly efficient rasterization. 3DGS-based methods~\cite{fisherrf,popgs,gauss-mi,activegs,activesplat,zhao2025self} adopt uncertainty or visibility-driven selection and show great promise for efficiency. However, current gradient-based algorithms assume surface-based occlusion and heavily rely on view-dependent color parameters. In contrast, X-ray imaging operates on a transmission model governed by the Beer-Lambert law~\cite{kak2001principles}, where projections serve as linear integrals of the density field without occlusion. This transmissive nature causes high spatial coupling among Gaussians along the ray, severely violating the diagonal approximation assumptions of gradient-based methods. Although uncertainty estimation and active acquisition have also been explored for CT implicit neural representations~\cite{vasconcelos2022uncertainr,zidane2025activenaf}, it remains challenging to radiative Gaussian Splatting. To address this gap, we abandon the gradient-based heuristic and introduce a forward parameter perturbation strategy specifically tailored to the explicit density parameters of radiative Gaussians.

\begin{figure}[t]
  \centering
  \includegraphics[width=\textwidth]{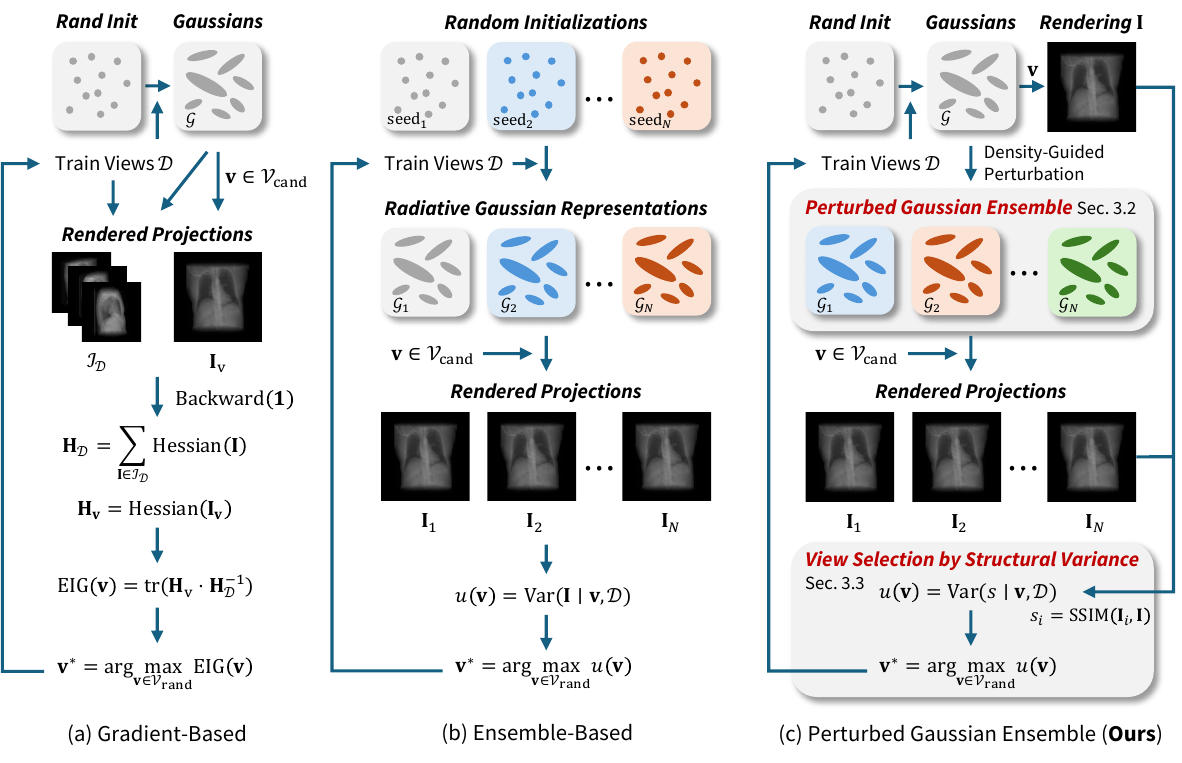}
  \caption{Comparison of active view selection paradigms for X-ray Gaussian Splatting. \textbf{(a)} Gradient-based method~\cite{fisherrf} estimates the expected information gain (EIG) of candidate views by computing a Fisher Information Matrix (FIM) via backpropagation, but suffers from gradient coupling and the absence of view-dependent parameters. \textbf{(b)} Ensemble-based method quantifies epistemic uncertainty by rendering disagreement across multiple Gaussian representations initialized with different random seeds, resulting in prohibitively high computational burden. \textbf{(c)} Our \emph{Perturbed Gaussian Ensemble} introduces density-guided parameter perturbations to efficiently construct an ensemble, wherein uncertain primitives exhibit pronounced behavioral randomness. By calculating the structural variance of the rendered projections, our approach achieves superior uncertainty modeling and enables more effective active view selection.} 
  \label{fig:method_overview}
\end{figure}

\section{Method}
\label{sec:method}
Given a sparse initial collection of measured X-ray projections $\mathcal{I}_\text{init}=\{\mathbf{I}_\text{init}^{(i)} \in \mathbb{R}^{H\times W}\}_{i=1,\dots,N_\text{init}}$, the aim is to iteratively select the next best view (NBV) from a pool of candidate scanner poses and incorporate the corresponding measurement into the training-view set until a predefined target number of observations $N_\text{target}$ is reached. Throughout this process, we progressively reconstruct the 3D volume to maximize the final reconstruction quality. To achieve this, we exploit the physical properties of radiative Gaussian Splatting to introduce a novel active view selection framework. As illustrated in \cref{fig:method_overview}, our approach models viewpoint uncertainty through the use of \emph{Perturbed Gaussian Ensemble}.

In this section, we present our proposed active view selection framework. First, we introduce the radiative Gaussian Splatting model and establish the theoretical basis in \cref{subsec:method_1}. Next, we detail the construction of \emph{Perturbed Gaussian Ensemble} through a density-guided parameter perturbation strategy in \cref{subsec:method_2}. Finally, \cref{subsec:method_3} formulates our view selection mechanism, which quantifies epistemic uncertainty by evaluating structural variance in the projection space.

\subsection{Preliminary: Radiative Gaussian Splatting}
\label{subsec:method_1}
To facilitate efficient and physically correct tomographic reconstruction, our active view selection mechanism is built on a radiative Gaussian Splatting framework, which reformulates 3D Gaussian Splatting (3DGS)~\cite{3dgs} for transmission imaging. 
Following~\cite{r2gaussian}, we model the object's 3D density field $\sigma$ as a mixture of $M$ radiative Gaussian kernels. Each kernel $G_i$ acts as a local volumetric density primitive parameterized by a central density scalar $\rho_i$, a mean position $\mathbf{p}_i \in \mathbb{R}^3$, and a 3D covariance matrix $\mathbf{\Sigma}_i \in \mathbb{R}^{3\times 3}$. The density field at a spatial location $\mathbf{x} \in \mathbb{R}^3$ is formulated as
\begin{equation}\label{eq:1}
    \sigma(\mathbf{x}) = \sum_{i=1}^M G_i^3(\mathbf{x}\mid \rho_i, \mathbf{p}_i, \mathbf{\Sigma}_i) = \sum_{i=1}^M \rho_i \exp\left(-\frac{1}{2}(\mathbf{x}-\mathbf{p}_i)^\top\mathbf{\Sigma}_i^{-1}(\mathbf{x}-\mathbf{p}_i)\right).
\end{equation}

The rendering process simulates the physical acquisition of X-ray projections. Due to the linearity of the integral operator, the total log-space projection is the sum of the line integrals of individual Gaussians:
\begin{equation}\label{eq:2}
    I_\text{r}(\mathbf{r}) = \sum_{i=1}^M G_i^2 (\hat{\mathbf{x}} \mid \rho_i\sqrt{\frac{2\pi\vert\mathbf{\Sigma}_i\vert}{\vert\hat{\mathbf{\Sigma}}_i\vert}}, \hat{\mathbf{p}}_i, \hat{\mathbf{\Sigma}}_i),
\end{equation}
where $I_\text{r}(\mathbf{r})$ is the rendered pixel value; $\hat{\mathbf{x}} \in \mathbb{R}^2$, $\hat{\mathbf{p}} \in \mathbb{R}^2$, $\hat{\mathbf{\Sigma}}\in\mathbb{R}^{2\times 2}$ are obtained by projecting the Gaussians onto the image plane from a viewpoint. The term $\sqrt{{2\pi\vert\mathbf{\Sigma}_i\vert}/{\vert\hat{\mathbf{\Sigma}}_i\vert}}$ (where $\vert\cdot\vert$ denotes the matrix determinant) acts as a normalization factor that ensures conservation of the integrated density mass, rendering the optimization of $\rho_i$ physically meaningful and view-consistent.

Radiative Gaussians are optimized by photometric L1 loss $\mathcal{L}_1$, D-SSIM loss $\mathcal{L}_\text{ssim}$~\cite{ssim}, and 3D total variation regularization $\mathcal{L}_\text{tv}$~\cite{tvloss}. The overall loss function is defined as
\begin{equation}
    \mathcal{L}=\mathcal{L}_1(\mathbf{I}_\text{r}, \mathbf{I}_\text{m}) + \lambda_1 \mathcal{L}_\text{ssim}(\mathbf{I}_\text{r}, \mathbf{I}_\text{m}) + \lambda_2 \mathcal{L}_\text{tv},
\end{equation}
where $\mathbf{I}_\text{r},\mathbf{I}_\text{m}$ are the rendered and measured projections. $\lambda_1,\lambda_2$ are weight terms.

\subsection{Perturbed Gaussian Ensemble}
\label{subsec:method_2}
Due to the transmissive and isotropic nature of X-rays, active view selection methods designed for natural-light scenes, such as FisherRF~\cite{fisherrf}, perform poorly in X-ray CT settings. To maintain real-time performance, FisherRF evaluates the expected information gain (EIG) using a diagonal approximation of the Fisher Information Matrix (FIM). While ignoring parameter correlations is valid for natural-light rendering, where pixels are dominated by a few front-most Gaussian primitives, this assumption fails in X-ray imaging. The transmissive nature of X-rays highly couples Gaussians along a ray, causing substantial mathematical bias in the EIG under a diagonal FIM.

To circumvent the computational intractability of evaluating a dense FIM and the inherent inaccuracies of its diagonal approximation, we abandon the gradient-based heuristic. Instead, we propose a forward, sampling-based approach that directly explores the coupled geometric uncertainty.

\subsubsection{Uncertainty Quantification via Rendering Disagreement.}
In sparse-view CT reconstruction, the lack of sufficient multi-view geometry constraints leads to severe ill-posedness, introducing significant uncertainty into the optimization of Gaussian fields. Let $\mathcal{D}$ denote the training data and $\theta$ represent the parameters of a radiative Gaussian Splatting model. For an arbitrary scanner pose $\mathbf{v}$, the predicted projection ${\mathbf{I}}(\mathbf{v})$ is obtained through the rendering process:
\begin{equation}
    {\mathbf{I}}(\mathbf{v}) = f(\mathbf{v};\theta).
\end{equation}
During optimization, we minimize the total loss to obtain the optimal model parameters:
\begin{equation}
    \hat{\theta} = \arg\min_\theta \mathcal{L}(\theta;\mathcal{D}) + \lambda R(\theta),
\end{equation}
where $R(\theta)$ represents the regularization term. Due to the highly non-convex nature of the objective function, the optimization problem admits multiple minima. Consequently, when initialized with different random seeds $\{k_i\}_{i=1,\dots,N}$, the model converges to distinct local optima:
\begin{equation}
    \hat{\theta}_i \sim q(\theta \mid \mathcal{D}),
\end{equation}
where $q(\theta \mid \mathcal{D})$ represents an empirical approximation of the true posterior $p(\theta \mid \mathcal{D})$. Thus, the posterior predictive distribution at viewpoint $\mathbf{v}$ can be approximated via Monte Carlo integration over the ensemble:
\begin{equation}
    p(\mathbf{I}(\mathbf{v}) \mid \mathcal{D}) \approx \frac{1}{N}\sum_{i=1}^N \delta \left({\mathbf{I}}(\mathbf{v}) - \mathbf{I}_i(\mathbf{v})\right),\quad \mathbf{I}_i(\mathbf{v}) = f(\mathbf{v};\hat{\theta}_i),
\end{equation}
where $\delta(\cdot)$ denotes the Dirac delta function. Therefore, the disagreement in the rendered projections across an ensemble of $N$ explicit Gaussian representations $\{\mathcal{G}_i\}_{i=1}^N$ initialized with different random seeds $\{k_i\}_{i=1,\dots,N}$ effectively captures the epistemic uncertainty at a given view. We quantify this uncertainty using the sample variance:
\begin{equation}\label{eq:var}
    \text{Var}\left[\mathbf{I}(\mathbf{v})\right] = \frac{1}{N-1} \sum_{i=1}^N \left(\mathbf{I}_i(\mathbf{v}) - \bar{\mathbf{I}}(\mathbf{v})\right)^2,\quad \bar{\mathbf{I}}(\mathbf{v}) = \frac{1}{N} \sum_{i=1}^N \mathbf{I}_i(\mathbf{v}).
\end{equation}

\subsubsection{Density-Guided Parameter Perturbation.}
Training an ensemble of Gaussian models is an intuitive and effective approach for viewpoint uncertainty modeling. However, this strategy requires training and storing $N$ independent Gaussian Splatting models, leading to a linear increase in computational and memory complexity, which is often prohibitive in practical applications. To address this, we propose \emph{Perturbed Gaussian Ensemble}, a simulation-based approach that leverages the intrinsic properties of the radiative Gaussian model.

Specifically, the variance in the rendered outputs across an ensemble at a given viewpoint is fundamentally induced by the randomness in the learned parameters of the individual Gaussian primitives. Therefore, a more efficient strategy is to train only a single Gaussian model $\mathcal{G} = \{G_i\}_{i=1,\dots,M}$ and, during uncertainty evaluation for candidate viewpoints, independently perturb the parameters of the Gaussian representation with repeated noise injections. This efficiently emulates an ensemble that would otherwise be obtained by training with different random seeds:
\begin{equation}
    \mathcal{G}_i=\{G_{i,j}\},\quad G_{i,j}=G_j\oplus\Delta_{i,j},\quad i=1,\dots,N,\quad j=1,\dots,M.
\end{equation}

To ensure that the rendering disagreement of this perturbed ensemble accurately reflects the epistemic uncertainty arising from the actual training process, the noise injected into each individual Gaussian primitive should be conditioned on its specific uncertainty level.

According to \cref{eq:1,eq:2}, the contribution of a Gaussian primitive during rendering is heavily determined by its density. High-density primitives typically correspond to real, well-defined solid structures (\eg, bones and dense organs). Because these structures strongly attenuate X-rays, they are usually well-constrained even under sparse-view settings. Consequently, the model exhibits high confidence in these regions.

In contrast, low-density primitives often lie near object boundaries (\eg, soft-tissue interfaces) or constitute background noise introduced to fit only a small number of rays, thereby exhibiting higher uncertainty. Moreover, due to insufficient multi-view constraints, the Gaussians tend to overfit the training views during optimization, giving rise to extremely elongated, needle-like artifacts or numerous tiny, low-density Gaussians distributed along the ray paths. These spurious components likewise exhibit low density and high uncertainty.

Motivated by this observation, we isolate a vulnerable Gaussian subset $\mathcal{G}_\text{low}$ comprising a specific fraction $\alpha$ (\eg, 10\%) of the total Gaussians in $\mathcal{G}$ with the lowest density values. We then stochastically perturb the density parameters of this subset to probe the stability of the reconstructed scene structure. Specifically, we generate an ensemble of $N$ independently perturbed models. For the $i$-th ensemble member $\mathcal{G}_i$ ($i = 1, \dots, N$), the density $\rho_{i,j}$ of a perturbed Gaussian primitive $G_{i,j}$ derived from $G_j \in \mathcal{G}$ ($j = 1, \dots, M$) is defined as:
\begin{equation}
    {\rho}_{i,j} = 
\begin{cases} 
\rho_j \cdot (1 + \epsilon_{i,j}), & \text{if } G_j \in \mathcal{G}_\text{low} \\
\rho_j, & \text{otherwise} 
\end{cases},
\end{equation}
where $\epsilon_{i,j}$ is a random scaling factor sampled from a uniform distribution:
\begin{equation}
    \epsilon_{i,j} \sim \text{Uniform}(-\beta, \beta), \quad\beta > 0.
\end{equation}
This targeted density scaling effectively injects variations into the most ambiguous regions of the volume while preserving the high-density, high-confidence anatomical structures.

\subsection{View Selection by Structural Variance}
\label{subsec:method_3}
The core objective of active view selection is to identify the candidate viewpoint that provides the most informative structural constraints to resolve existing geometric ambiguities. For an arbitrary candidate viewpoint $\mathbf{v}$, rendered projections are generated using the \emph{Perturbed Gaussian Ensemble} $\{\mathcal{G}, \mathcal{G}_1, \dots, \mathcal{G}_N\}$ via the rendering function:
\begin{equation}
    \mathbf{I}(\mathbf{v}) = f\left(\mathbf{v};\theta(\mathcal{G})\right),\quad \mathbf{I}_i(\mathbf{v}) = f\left(\mathbf{v};\theta(\mathcal{G}_i)\right),\quad i=1,\dots,N.
\end{equation}
To quantify the epistemic uncertainty at viewpoint $\mathbf{v}$, we leverage the Structural Similarity Index Measure (SSIM)~\cite{ssim} in the projection space. This allows us to evaluate the macroscopic structural disagreement induced by our parameter perturbations:
\begin{equation}
    s_i(\mathbf{v}) = \text{SSIM}\left(\mathbf{I}(\mathbf{v}),\mathbf{I}_i(\mathbf{v})\right).
\end{equation}
The uncertainty score $u(\mathbf{v})$ for the candidate viewpoint is formulated as the sample variance of these $N$ structural similarity scores:
\begin{equation}
    u(\mathbf{v}) = \frac{1}{N-1} \sum_{i=1}^{N} \left( s_i(\mathbf{v}) - \bar{s}(\mathbf{v}) \right)^2,
\end{equation}
where $\bar{s}(\mathbf{v})$ denotes the mean of the $N$ SSIM scores.

At each view selection iteration $t$, a new perturbed ensemble is generated from the most recently updated Gaussian representation $\mathcal{G}^{(t)}$. For every viewpoint from the candidate pool $\mathcal{V}_\text{cand}^{(t)}$, we calculate the epistemic uncertainty score using this ensemble. Finally, the viewpoint exhibiting the highest uncertainty is selected as the next best view:
\begin{equation}
    \mathbf{v}^{*(t)} = \arg \max_{\mathbf{v}\in \mathcal{V}_\text{cand}^{(t)}} u(\mathbf{v}).
\end{equation}
The corresponding physical measurement from $\mathbf{v}^{*(t)}$ is then acquired and appended to the training set for the subsequent phase of progressive optimization.

\begin{table*}[t]

\caption{Quantitative comparisons of 3D tomographic reconstruction and novel view synthesis across different view selection strategies under two protocols. Performance is evaluated in terms of 3D PSNR$\uparrow$ (PSNR$_3$, dB), 3D SSIM$\uparrow$ (SSIM$_3$), 2D PSNR$\uparrow$ (PSNR, dB), and 2D SSIM$\uparrow$ (SSIM). The \textbf{best} and \underline{second-best} results are highlighted. \emph{Random}, \emph{Uniform}, and \emph{FPS} denote random view selection, uniform sampling, and Farthest Point Sampling, respectively.}
\label{tab:comp_3d}

\scriptsize
\centering
\renewcommand\arraystretch{1.2}
\setlength{\tabcolsep}{2pt}

\begin{tabularx}{0.95\textwidth}{c|c|>{\centering\arraybackslash}X>{\centering\arraybackslash}X>{\centering\arraybackslash}X>{\centering\arraybackslash}X|>{\centering\arraybackslash}X>{\centering\arraybackslash}X>{\centering\arraybackslash}X>{\centering\arraybackslash}X}

\multirow{2.5}{*}{\textbf{Category}} &
\multirow{2.5}{*}{\textbf{Method}} &
\multicolumn{4}{c}{\textbf{24-view}} &
\multicolumn{4}{c}{\textbf{36-view}}\\

\cmidrule{3-6}\cmidrule{7-10}

&& \textbf{PSNR$_3$} & \textbf{SSIM$_3$}
& \textbf{PSNR} & \textbf{SSIM}
& \textbf{PSNR$_3$} & \textbf{SSIM$_3$}
& \textbf{PSNR} & \textbf{SSIM}\\

\hline\hline

\multicolumn{10}{c}{\textbf{Synthetic Dataset}}\\
\hline

\multirow{3}{*}{Rule-based}
& Random & 32.629 & 0.881 & 41.862 & 0.954 & 34.823 & 0.915 & 44.552 & 0.961\\
& Uniform & \underline{33.562} & 0.890 & 43.260 & 0.961 & \underline{35.877} & \underline{0.921} & 46.053 & 0.968\\
& FPS & {33.508} & \underline{0.891} & 43.081 & 0.957 & 35.367 & 0.919 & 45.386 & 0.964\\

\hline

\multirow{3}{*}{2D-based}
& TOPIQ~\cite{topiq} & 33.000 & 0.882 & 42.898 & 0.961 & 34.951 & 0.912 & 45.483 & 0.967\\
& MANIQA~\cite{maniqa} & 32.319 & 0.882 & 41.482 & 0.953 & 34.543 & 0.913 & 44.516 & 0.962\\
& MUSIQ~\cite{musiq} & 33.083 & 0.886 & 42.675 & 0.957 & 35.030 & 0.916 & 45.133 & 0.965\\

\hline

\multirow{2}{*}{3D-based}
& FisherRF~\cite{fisherrf} & 33.347 & 0.887 & \underline{43.289} & \underline{0.962} & {35.551} & {0.919} & \underline{46.212} & \underline{0.969}\\
& {Ours} & \textbf{34.078} & \textbf{0.896} & \textbf{44.069} & \textbf{0.966} & \textbf{36.226} & \textbf{0.926} & \textbf{46.849} & \textbf{0.971}\\

\hline

\multicolumn{10}{c}{\textbf{Real-World Dataset}}\\
\hline

\multirow{2}{*}{Rule-based}
& Random & 36.112 & 0.903 & 42.992 & 0.985 & 36.765 & 0.921 & 44.965 & 0.989\\
& FPS & 36.134 & 0.905 & 43.073 & 0.985 & 36.898 & 0.925 & 45.171 & 0.989\\

\hline

\multirow{3}{*}{2D-based}
& TOPIQ~\cite{topiq} & 36.208 & 0.906 & 43.121 & 0.986 & 37.167 & 0.928 & 45.644 & 0.990\\
& MANIQA~\cite{maniqa} & 36.222 & 0.906 & \underline{43.476} & \textbf{0.986} & 37.187 & \underline{0.929} & 45.644 & \underline{0.990}\\
& MUSIQ~\cite{musiq} & \underline{36.257} & \underline{0.906} & 43.293 & 0.985 & 37.119 & 0.925 & 45.494 & 0.990\\

\hline

\multirow{2}{*}{3D-based}
& FisherRF~\cite{fisherrf} & 36.205 & 0.902 & 43.179 & 0.985 & \underline{37.258} & 0.926 & \underline{46.022} & 0.990\\
& {Ours} & \textbf{36.399} & \textbf{0.909} & \textbf{43.546} & \underline{0.986} & \textbf{37.480} & \textbf{0.932} & \textbf{46.046} & \textbf{0.991}\\

\hline

\end{tabularx}
\end{table*}

\begin{figure}[t]
  \centering
  \includegraphics[width=\textwidth]{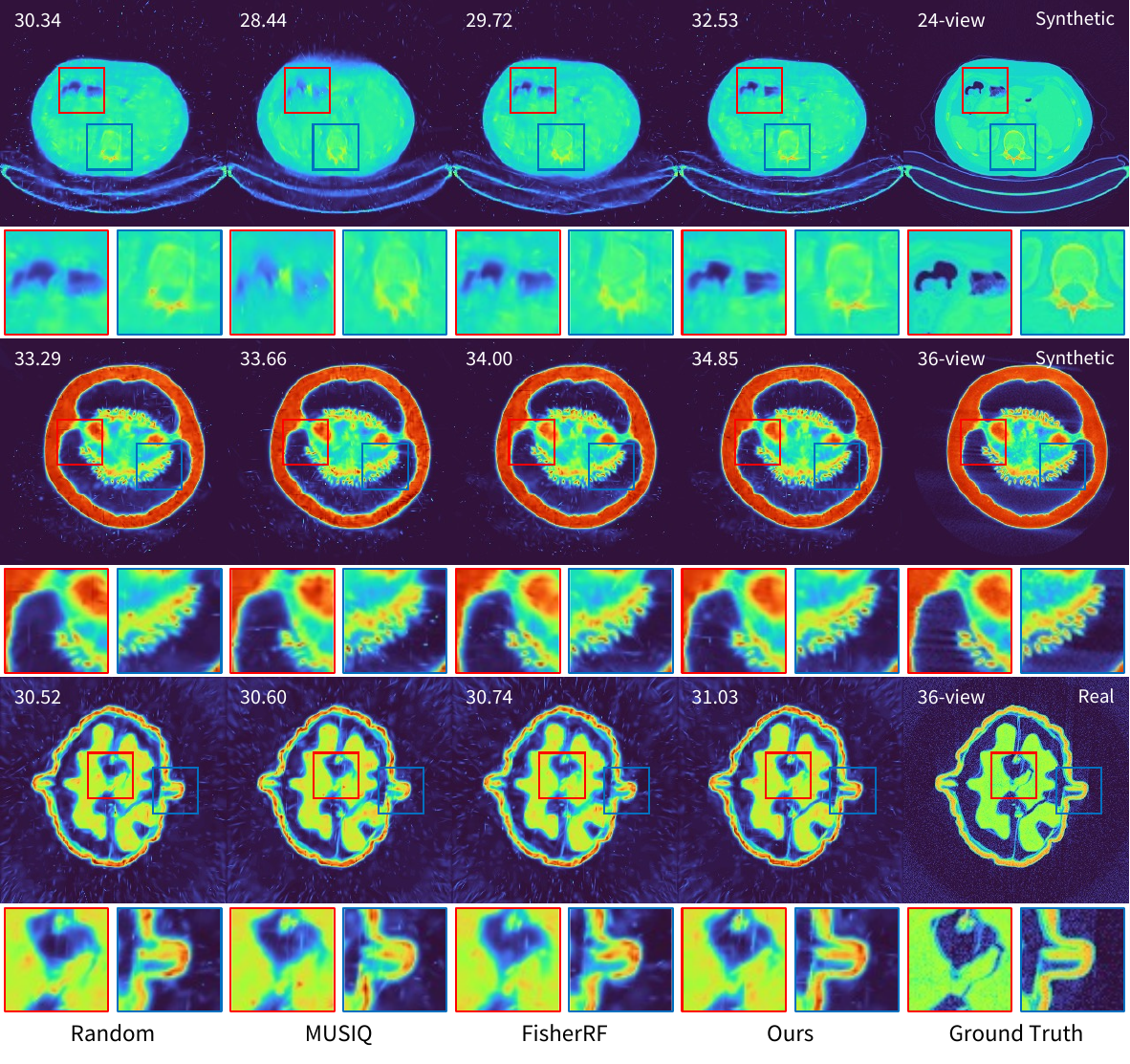}
  \caption{Visual comparisons of the reconstructed 3D volumes using different view selection strategies. The 3D PSNR$\uparrow$ (dB) for each scene is displayed at the top-left corner of the corresponding image. Our approach consistently achieves the highest reconstruction quality and better preserves fine structural details.}
  \label{fig:comp1}
\end{figure}

\section{Experiments}
\label{sec:exp}

\subsection{Experimental Settings}
\subsubsection{Datasets.}
Following the experimental settings of R$^2$-Gaussian~\cite{r2gaussian}, we evaluate our method on both synthetic and real-world datasets. The synthetic benchmark comprises 15 CT volumes aggregated from various public datasets~\cite{lidc-idri,pancreasct,xplantdata,scivisdata}. For real-world evaluation, we utilize three cases from the FIPS dataset~\cite{FIPS_CT_dataset}, reconstructing pseudo-ground-truth (pseudo-GT) volumes via FDK~\cite{fdk}. We adopt an object-centric hemispherical acquisition trajectory rather than a conventional circular orbit. The candidate viewpoint pool for active selection consists of 448 poses uniformly sampled across two concentric hemispheres at different radii, with all optical axes oriented toward the volume center. For novel view synthesis evaluation, 103 test viewpoints are randomly sampled from a single hemisphere per scene. All GT and pseudo-GT projections are simulated using DiffDRR~\cite{diffdrr}.

\subsubsection{Protocols.}
We initialize the Gaussian primitives with 50k random points and train the radiative GS model for 30k iterations per scene under the default configuration. Following FisherRF~\cite{fisherrf}, active view selection is performed at progressively increasing iteration intervals. To ensure stable optimization, we schedule the final view selection prior to the termination of the GS densification phase. At each selection step, a single optimal view is acquired and appended to the training set. We evaluate our method and all baselines under two final view-budget protocols: $N_\text{target}=24$ and 36 (inclusive of two initial views).

\begin{table}[t]
\caption{Quantitative comparisons of 3D tomographic reconstruction on sparser-view protocols. Performance is evaluated in terms of 3D PSNR$\uparrow$ (PSNR$_3$, dB) and 3D SSIM$\uparrow$ (SSIM$_3$). The \textbf{best} and \underline{second-best} results are highlighted.}
\label{tab:comp_2d}

\scriptsize
\centering
\renewcommand\arraystretch{1.2}
\setlength{\tabcolsep}{2pt}

\begin{tabularx}{0.95\linewidth}{c|>{\centering\arraybackslash}X>{\centering\arraybackslash}X|>{\centering\arraybackslash}X>{\centering\arraybackslash}X|>{\centering\arraybackslash}X>{\centering\arraybackslash}X|>{\centering\arraybackslash}X>{\centering\arraybackslash}X}

\multirow{2.5}{*}{\textbf{Method}} &
\multicolumn{2}{c}{\textbf{6-view}} &
\multicolumn{2}{c}{\textbf{8-view}} &
\multicolumn{2}{c}{\textbf{12-view}} &
\multicolumn{2}{c}{\textbf{16-view}}\\

\cmidrule{2-3}\cmidrule{4-5}\cmidrule{6-7}\cmidrule{8-9}

& \textbf{PSNR$_3$} & \textbf{SSIM$_3$}
& \textbf{PSNR$_3$} & \textbf{SSIM$_3$}
& \textbf{PSNR$_3$} & \textbf{SSIM$_3$}
& \textbf{PSNR$_3$} & \textbf{SSIM$_3$}\\

\hline\hline

Random & 24.58 & 0.711 & 25.14 & 0.729 & 28.54 & 0.805 & 29.98 & 0.836\\
FPS & 25.29 & 0.742 & 26.85 & \underline{0.780} & 29.10 & 0.822 & 30.50 & 0.846\\

Uniform & 23.77 & 0.697 & 26.84 & 0.772 & 28.95 & 0.811 & \underline{31.16} & \underline{0.851}\\
MUSIQ~\cite{musiq} & 25.54 & 0.741 & 27.04 & 0.767 & 29.86 & \underline{0.824} & 31.03 & 0.847\\

FisherRF~\cite{fisherrf} & \underline{25.80} & \underline{0.748} & \underline{27.43} & 0.779 & \underline{29.87} & 0.823 & 30.79 & 0.846\\
{Ours} & \textbf{25.93} & \textbf{0.758} & \textbf{27.90} & \textbf{0.786} & \textbf{30.29} & \textbf{0.830} & \textbf{31.30} & \textbf{0.855}\\

\hline

\end{tabularx}
\end{table}

\begin{figure}[t]
  \centering
  \includegraphics[width=\textwidth]{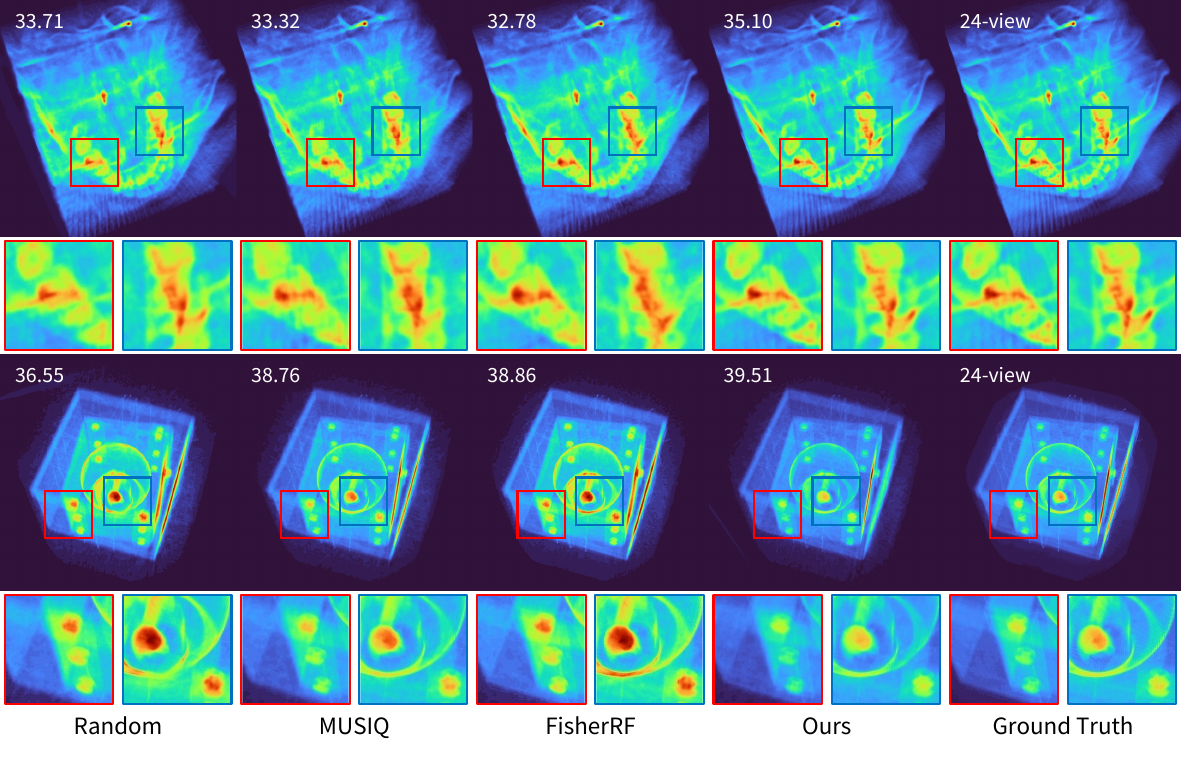}
  \caption{Visual comparisons of novel view synthesis results across different view selection strategies. The PSNR$\uparrow$ (dB) for each scene is displayed at the top-left corner of each image. Our approach achieves the highest rendering quality.}
  \label{fig:comp2}
\end{figure}

\subsubsection{Baselines.}
Previous active view selection strategies generally fall into two paradigms: 2D-based approaches that evaluate rendered image quality, and 3D-based approaches that leverage model uncertainty. We benchmark our method against state-of-the-art baselines from both categories, alongside three standard heuristics: random selection, uniform sampling, and Farthest Point Sampling (FPS). \textbf{i.} 2D-based: We adopt three no-reference image quality assessment (IQA) metrics, MUSIQ~\cite{musiq}, MANIQA~\cite{maniqa}, and TOPIQ~\cite{topiq}, implemented via the PyIQA~\cite{pyiqa} toolbox. \textbf{ii.} 3D-based: We compare against the prevailing uncertainty-driven baseline, FisherRF~\cite{fisherrf}. All baselines are re-implemented upon the radiative GS framework.

\subsubsection{Implementation Details.}
Our pipeline is implemented in PyTorch~\cite{pytorch} and optimized via Adam~\cite{adam}. We set the ensemble size $N=10$, perturbation parameters $\alpha=10\%$ and $\beta=0.5$, and adopt the learning rates and loss weights from R$^2$-Gaussian~\cite{r2gaussian}. All experiments are conducted on a single NVIDIA A40 GPU. For quantitative evaluation, we report the Peak Signal-to-Noise Ratio (PSNR) computed over the full 3D volume, and the Structural Similarity Index Measure (SSIM)~\cite{ssim} averaged across 2D slices in the axial, coronal, and sagittal planes.

\subsection{Results}

\subsubsection{Tomographic Reconstruction.}
\Cref{tab:comp_3d} presents the quantitative results of 3D tomographic reconstruction across various active view selection strategies. Our approach demonstrates superior performance in both synthetic and real-world scenarios. In terms of 3D PSNR, our method surpasses the second-best baseline by margins of up to 0.68 dB on synthetic data and 0.22 dB on real-world data. Notably, FisherRF underperforms FPS under the synthetic 24-view setting. This degradation stems from FisherRF’s diagonal approximation of the Fisher Information Matrix. In X-ray imaging, strongly coupled Gaussians along projection rays violate this assumption, yielding severely biased information gain estimates. Qualitatively, as illustrated in \cref{fig:comp1}, our method yields the most visually accurate reconstructions, faithfully preserving fine structural details, significantly suppressing boundary artifacts and background noise. We also evaluated our approach on sparser-view settings (6/8/12/16 views). As shown in \cref{tab:comp_2d}, ours consistently maintains superior performance.

\subsubsection{Novel View Synthesis.}
We also evaluate our method's performance on novel view synthesis using the synthetic dataset. The qualitative comparisons and quantitative results are reported in \cref{fig:comp2} and \cref{tab:comp_3d}, respectively. Our approach achieves the best overall rendering quality under both protocols, demonstrating an improvement of up to 0.78 dB in PSNR. Visual comparisons indicate that our method better preserves local structural details, particularly in high-density areas. This improvement stems from our method's ability to acquire more informative viewpoints, which effectively mitigates the needle-like artifacts and noise that typically appear along rays passing through dense regions, \eg, bones.

\subsection{Ablation Study and Analysis}

\subsubsection{Component Analysis.}
To demonstrate the impact of our proposed structural variance-based view selection mechanism, we replace SSIM with L1 error and PSNR, respectively, and evaluate the performance under the 24-view protocol. As shown in the first section of \cref{tab:abla}, switching to L1 error or PSNR as the uncertainty metric leads to a significant performance drop. There are two primary reasons for this degradation. Firstly, L1 error and PSNR are pixel-wise, independently computed metrics based on absolute or mean-squared errors. They overlook the spatial arrangement and inter-dependencies of adjacent pixels, making them less sensitive to high-frequency topological changes, such as the structural discontinuities caused by perturbing needle-like artifacts. Secondly, due to the linear integral nature of X-ray imaging, perturbing the density of Gaussian primitives inevitably causes overall intensity shifts in the projection space. L1 and PSNR are highly susceptible to these absolute luminance fluctuations, which can easily overwhelm the subtle variance signals originating from genuine geometric ambiguities. In contrast, SSIM inherently incorporates luminance and contrast normalization, effectively decoupling absolute intensity shifts from structural information. This enables the SSIM variance to act as a more robust and precise indicator of true epistemic uncertainty and geometric disagreement.

\begin{table}[t]
\caption{Ablation study results on the uncertainty quantification metric, ensemble size ($N$), perturbed ratio ($\alpha$), and density scaling amplitude ($\beta$). Performance is evaluated in terms of 3D PSNR$\uparrow$ (PSNR$_3$, dB) and 3D SSIM$\uparrow$ (SSIM$_3$). Within each section, the \textbf{best} and \underline{second-best} results are highlighted. $^\dagger$ indicates the default setting.}
\label{tab:abla}

\scriptsize
\centering
\renewcommand\arraystretch{1.2}
\setlength{\tabcolsep}{2pt}

\begin{tabularx}{0.95\linewidth}{>{\centering\arraybackslash}X>{\centering\arraybackslash}X>{\centering\arraybackslash}X|>{\centering\arraybackslash}X>{\centering\arraybackslash}X>{\centering\arraybackslash}X}

\textbf{Setting} & \textbf{PSNR$_3\uparrow$} & \textbf{SSIM$_3\uparrow$} & \textbf{Setting} & \textbf{PSNR$_3\uparrow$} & \textbf{SSIM$_3\uparrow$}\\

\hline\hline

L1 & \underline{33.644} & \underline{0.894} & $\alpha=5\%$ & 33.680 & 0.893\\
PSNR & 33.390 & 0.889 & $\alpha=10\%^\dagger$ & \textbf{34.078} & \underline{0.896}\\
SSIM$^\dagger$ & \textbf{34.078} & \textbf{0.896} & $\alpha=15\%$ & \underline{33.938} & \textbf{0.896}\\

\cline{1-3}

$N=5$ & {33.952} & {0.895} & $\alpha=20\%$ & 33.589 & 0.892\\

\cline{4-6}

$N=10^\dagger$ & \textbf{34.078} & \underline{0.896} & $\beta=0.1$ & 34.040 & \underline{0.896}\\
$N=15$ & \underline{34.034} & \textbf{0.896} & $\beta=0.3$ & \underline{34.057} & \textbf{0.897}\\
$N=20$ & 33.865 & 0.895 & $\beta=0.5^\dagger$ & \textbf{34.078} & 0.896\\
$N=40$ & 33.670 & 0.893 & $\beta=1.0$ & 33.338 & 0.890\\

\hline

\end{tabularx}
\end{table}

\subsubsection{Parameter Analysis.}
We evaluate the impact of ensemble size ($N$), ratio of perturbed Gaussian primitives ($\alpha$), and density scaling amplitude ($\beta$) on reconstruction quality, as detailed in the last three sections of \cref{tab:abla}. Empirically, we observe that an ensemble size of $N=10$ or $15$ yields the optimal trade-off. Increasing $N$ to 40 leads to a performance degradation. We attribute this to the over-smoothing of the structural disagreement signal: A larger sample size dilutes the impact of extreme catastrophic structural failures caused by perturbations, thereby reducing the discriminative contrast of the SSIM variance across views. A moderate size $N$ successfully preserves the sharpness of the uncertainty landscape, allowing the active agent to accurately penalize geometric degeneracies.

The ratio $\alpha$ controls the proportion of Gaussians subjected to stochastic scaling, with $10\%$ yielding the best results. Both excessively low and high ratios deteriorate the performance. When $\alpha$ is too small, the perturbation is overly restricted to background noise or void space, failing to include the primitives that constitute geometric degeneracies, resulting in a flat structural disagreement landscape. Conversely, a higher ratio indiscriminately extends the perturbations into well-constrained, high-confidence anatomical solids (\eg, bones). Due to the linear integral nature of X-ray projection, modulating these high-density structures induces massive, non-informative global variations that mask the subtle uncertainty signals from ambiguous regions.

For the scaling amplitude $\beta$, where the density scaling factor is sampled from $\text{Uniform}(-\beta, \beta)$, both overly conservative and excessively aggressive perturbations degrade the final reconstruction quality. A small $\beta$ injects insufficient noise to disrupt the fragile geometric equilibrium of overfitting artifacts, providing weak guidance for view selection. Conversely, an excessively large $\beta$ introduces severe out-of-distribution intensity shifts that completely distort local geometry. This structural collapse causes ubiquitous SSIM degradation across all views, diminishing the relative variance contrast and blinding the acquisition function.

\begin{table}[t]
\caption{Ablation study results on different perturbations. Performance is evaluated in terms of 3D PSNR$\uparrow$ (PSNR$_3$, dB) and 3D SSIM$\uparrow$ (SSIM$_3$). The \textbf{best} and \underline{second-best} results are highlighted. $^\dagger$ indicates our approach.}
\label{tab:perturb}

\scriptsize
\centering
\renewcommand\arraystretch{1.2}
\setlength{\tabcolsep}{1pt}

\begin{tabularx}{0.95\linewidth}{>{\centering\arraybackslash}X|>{\centering\arraybackslash}X>{\centering\arraybackslash}X|>{\centering\arraybackslash}X>{\centering\arraybackslash}X|>{\centering\arraybackslash}X>{\centering\arraybackslash}X|>{\centering\arraybackslash}X>{\centering\arraybackslash}X}

\multirow{2.5}{*}{\textbf{Views}} & \multicolumn{2}{c}{\textbf{Dropout}} & \multicolumn{2}{c}{\textbf{Position Jittering}} & \multicolumn{2}{c}{\textbf{PJ + SS}} & \multicolumn{2}{c}{\textbf{Stochastic Scaling}$^\dagger$} \\

\cmidrule{2-3}\cmidrule{4-5}\cmidrule{6-7}\cmidrule{8-9}

& \textbf{PSNR$_3$} & \textbf{SSIM$_3$} &
\textbf{PSNR$_3$} & \textbf{SSIM$_3$} &
\textbf{PSNR$_3$} & \textbf{SSIM$_3$} &
\textbf{PSNR$_3$} & \textbf{SSIM$_3$}\\

\hline\hline

\textbf{24} & 32.718 & 0.883 & \underline{34.061} & 0.895 & 33.886 & \underline{0.896} & \textbf{34.078} & \textbf{0.896}\\
\textbf{36} & 34.607 & 0.924 & 36.112 & 0.925 & \underline{36.200} & \underline{0.925} & \textbf{36.226} & \textbf{0.926}\\

\hline

\end{tabularx}
\end{table}

\subsubsection{Perturbation Strategy Analysis.}
To demonstrate the effectiveness of our perturbation strategy (Stochastic Scaling), we further compare it with several alternative perturbations. As shown in \cref{tab:perturb}, our method achieves the best results. Dropout and Position Jittering tend to mislead the estimation of uncertain regions, thus leading to suboptimal view selections.

\section{Conclusion}
We present a novel active view selection framework for sparse-view CT reconstruction utilizing radiative Gaussian Splatting. To overcome the theoretical flaws of existing active learning in X-ray imaging, we introduce a physics-aware active view selection method based on \emph{Perturbed Gaussian Ensemble}. By injecting stochastic density perturbations into under-constrained, low-density primitives, our approach exposes structural vulnerabilities quantified via projection-space SSIM variance. This mechanism successfully targets and suppresses artifacts. Extensive evaluations confirm that our method yields superior volumetric reconstructions compared to existing baselines. This work bridges the gap between active learning and explicit radiative fields, advancing the practical deployment of 3DGS in dose-sensitive clinical and industrial settings.

\section*{Acknowledgements}

This research is supported by the National Artificial Intelligence Research Resource (NAIRR) Pilot Awards NAIRR250199, NAIRR260019, and NAIRR260077, the AMD University Program’s AI \& HPC Cluster, NVIDIA Academic Grant Program, and Lambda's Research Grant. Computational resources are also provided by Delta and DeltaAI at the National Center for Supercomputing Applications (NCSA) through ACCESS allocations CIS250012, CIS250816, and CIS251188. 

\bibliographystyle{splncs04}
\bibliography{main}
\end{document}